\documentclass[letterpaper,journal]{IEEEtran}
\IEEEoverridecommandlockouts                              

\usepackage{amssymb}  
\usepackage{bm}
\usepackage{graphicx}
\usepackage[table]{xcolor} 
\usepackage{booktabs}
\usepackage{amsmath,amsfonts}
\usepackage[ruled,vlined]{algorithm2e}
\usepackage{array}
\usepackage[caption=false,font=normalsize,labelfont=sf,textfont=sf]{subfig}
\usepackage{textcomp}
\usepackage{stfloats}
\usepackage{url}
\usepackage{verbatim}
\usepackage{graphicx}
\usepackage{cite}
\usepackage{hyperref}

\hypersetup{
    colorlinks=true,
    linkcolor=blue,
    filecolor=magenta,  
    citecolor=magenta,      
    urlcolor=blue,
}

\SetCommentSty{mycommfont}
\SetAlgoSkip{}
\DontPrintSemicolon  

\title{\LARGE \bf
Generative Models from and for Sampling-Based MPC: \\
A Bootstrapped Approach for \\
Adaptive Contact-Rich Manipulation
}

\author{
Lara Brudermüller$^{1,2,*}$, 
Brandon Hung$^{2}$, 
Xinghao Zhu$^{2}$, 
Jiuguang Wang$^{2}$, \\
Nick Hawes$^{1}$, 
Preston Culbertson$^{2,3,\dagger}$, 
Simon Le Cleac'h$^{2,\dagger}$%
\thanks{$^{1}$Oxford Robotics Institute, University of Oxford, UK.}%
\thanks{$^{2}$Robotics and AI Institute (RAI), Boston, USA.}%
\thanks{$^{3}$Cornell University, Ithaca, NY, USA.}%
\thanks{$^{*}$This work was conducted while LB was an intern at RAI.}%
\thanks{$^{\dagger}$PC and SLC advised this work equally.}%
}

\begin{document}

\maketitle
\thispagestyle{empty}
\pagestyle{empty}

\begin{abstract}
We present a generative predictive control (GPC) framework that amortizes sampling-based Model Predictive Control (SPC) by bootstrapping it with conditional flow-matching models trained on SPC control sequences collected in simulation. Unlike prior work relying on iterative refinement or gradient-based solvers, we show that meaningful proposal distributions can be learned directly from noisy SPC data, enabling more efficient and informed sampling during online planning. We further demonstrate, for the first time, the application of this approach to real-world contact-rich loco-manipulation with a quadruped robot. Extensive experiments in simulation and on hardware show that our method improves sample efficiency, reduces planning horizon requirements, and generalizes robustly across task variations.
\end{abstract}
\vspace{-7pt}


\section{Introduction}

Reactive contact-rich (loco-)manipulation in high-dimensional state and action spaces poses significant challenges for real-time control. Sampling-based Model Predictive Control (SPC) offers a principled framework to address these challenges by solving trajectory optimization problems online with a model in the loop, enabling adaptive behavior and constraint satisfaction \cite{li2024drop, howell2022predictive, jankowski2023vp, bhardwaj2022storm}. However, the computational cost of forward simulation, combined with the challenge of effectively exploring the search space in high-dimensional, contact-rich environments, limits the applicability of real-time optimization to more complex behaviors and higher-frequency control.

A promising line of work seeks to amortize the computational burden of online optimization by shifting it to an offline phase~\cite{ichter2018learning, fishman2023motion, zhou2024diffusion}. The key idea is to collect high quality data and train a generative model to capture a distribution of useful actions or control sequences. At test time, this model can be used to guide or warmstart the sampling distribution. The method attempts to drastically improve solution quality and efficiency by focusing sampling on high-likelihood, constraint-satisfying regions of the action space.

Recent advances in generative modeling, particularly diffusion and flow-matching models, have shown strong performance in learning expressive end-to-end policies for dexterous manipulation tasks \cite{black2410pi0, chi2023diffusion, zhao2023learning}. Offline model-based reinforcement learning methods \cite{hansen2023td, dadiotis2025dynamic} also show strong performance approximating optimal solutions by leveraging precomputed data to enable fast runtime control via policy networks. However, these methods are often limited by the scope of their immense training data and struggle to generalize to out-of-distribution (OOD) states or tasks.
In response to these limitations, several recent works demonstrate that bootstrapping online planners with offline-trained generative models leads to faster convergence, better exploration, and more robust performance in complex environments \cite{zhou2024diffusion, qi2025strengthening, kurtz2025generative, wang2025bootstrapped}.
In this paper, we focus on how offline data collection and generative modeling can both accelerate and guide online sampling-based MPC in contact-rich, high-dimensional settings while maintaining the flexibility and adaptability of online optimization.

\paragraph*{Contributions}
We propose a \textit{generative predictive control (GPC) framework} that \textit{bootstraps} SPC with conditional flow-matching models trained on SPC control sequences collected in simulation. To the best of our knowledge, we are the first to show that meaningful \textit{proposal distributions can be learned directly from noisy SPC data} without requiring expert refinement or numerical solvers. We are also the first to demonstrate that this approach improves sample efficiency and \textit{generalizes robustly to task variations} in both simulation and\textit{real hardware in a contact-rich loco-manipulation task}.

\begin{figure}[t]
\centering
\includegraphics[width=\linewidth]{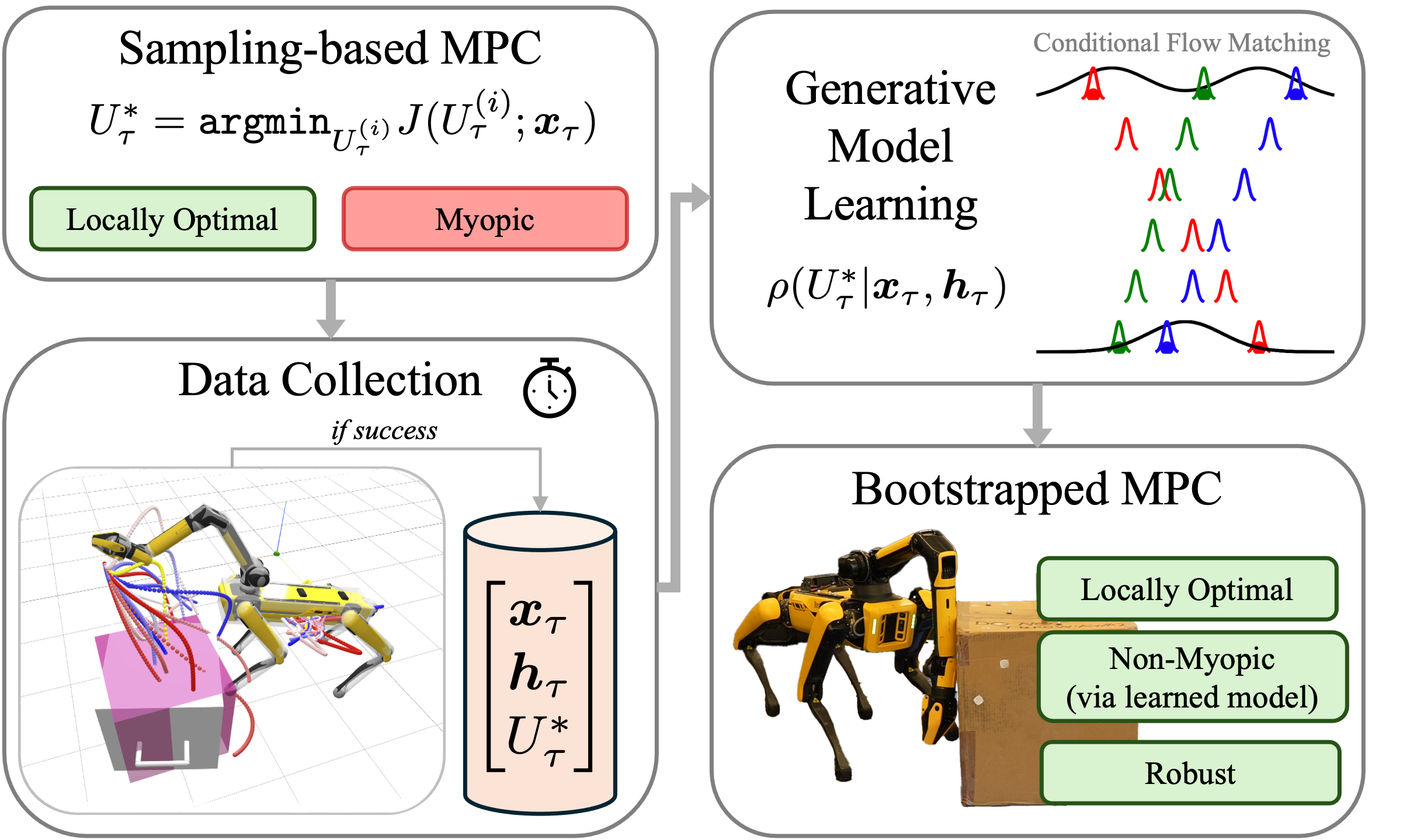}
\caption{\textit{Generative predictive control (GPC) framework for bootstrapping sampling-based MPC}. We collect open-loop control sequences from an SPC algorithm in simulation and use them to train a generative proposal distribution. At test time, this model guides and amortizes online MPC, enabling non-myopic, constraint-satisfying behavior with improved sample efficiency and robustness in contact-rich, high-dimensional settings.}
\vspace{-10pt}
\label{fig:teaser}
\end{figure}


\section{Related Work}
\label{sec:related_work}
\paragraph*{SPC for Contact-Rich Manipulation}
SPC has been widely adopted for its robustness to nonconvex and discontinuous problems, particularly in contact-rich robotic tasks \cite{williams2017model, kobilarov2012cross, howell2022predictive, li2024drop, jankowski2023vp, xue2024full, pan2024model}. These methods typically optimize a trajectory distribution by iteratively sampling candidate controls and selecting actions based on forward-simulated costs. Their performance is often limited by the computational cost of forward dynamics, especially when simulating contact interactions or systems with many degrees of freedom (DOF). 
While prior work has sought to speed up forward simulation, e.g., via quasi-static approximations~\cite{pang2021convex} or learned dynamics models~\cite{jain2025smooth}, these approaches often trade off fidelity or depend on highly accurate model learning.
In contrast, we do not aim to replace the dynamics model but rather to amortize the trajectory optimization itself.
We do this by learning generative models over control distributions derived from open-loop control sequences that either led to task success or incurred low cost in offline sampling-based MPC, enabling faster and more informed sampling at test time.

\paragraph*{Amortizing Online Optimization via Offline Learning}
Recent work leverages offline data to reduce the computational burden of control algorithms at runtime. Common approaches include behavior cloning on expert demonstrations and planner rollouts \cite{black2410pi0, chi2023diffusion, zhao2023learning} or model-based RL \cite{hansen2023td, dadiotis2025dynamic}.
When sourced from high-quality planners (such as sampling-based MPC), this data enables training generative models that can guide or initialize online control.
Several works have explored learning from planner-generated trajectories to approximate optimal solutions and accelerate planning \cite{dalal2024neural, carvalho2023motion, fishman2023motion, urain2022learning}. This idea underpins Approximate MPC (AMPC), where learned models bootstrap or replace expensive solvers \cite{carius2020mpc}. A representative example \cite{huang2024toward} uses diffusion models to approximate near-globally optimal MPC solutions from locally optimal trajectories generated by a numerical solver \cite{wachter2006implementation}. Yet, the learned models are not used to guide sampling but rather to replace the solver entirely.
%
Our work is most closely related to~\cite{kurtz2025generative}, which also bootstraps SPC using generative models trained on SPC control sequences. 
Their method alternates online data collection with model updates, but we find this iterative refinement can collapse the multi-modal control distribution important for effective sampling.
In contrast, our method trains a generative model over \emph{offline} data (noisy SPC rollouts) and achieves strong performance, showing that bootstrapped SPC does not require iterative refinement. To the best of our knowledge, both~\cite{kurtz2025generative} and our approach are the first to learn from SPC data rather than trajectories from gradient-based solvers.

\paragraph*{Learning Sampling Distributions for Online MPC}
Another line of work aims to improve SPC by learning structured priors over control sequences in latent action spaces using generative models \cite{sacks2023learning, power2024learning}. 
These methods typically rely on expert demonstrations, which lack exploratory diversity and require complex bi-level training to learn both the latent spaces and their distributions. In contrast, we focus on directly leveraging the diverse data produced by sampling-based MPC during offline data collection. Freed from real-time constraints, we can instead expand the search space during planning to yield richer control sequences that support efficient sampling through simpler generative models.

\paragraph*{Infinite-Horizon Value Approximation and MPC}
A complementary line of research seeks to reduce the myopia of finite-horizon MPC by learning \emph{infinite-horizon value functions} and integrating them into the control loop~\cite{jordana2025infinite, hoeller2020deep, lowrey2018plan, hatch2021value}. These methods approximate an infinite-horizon value signal over states, which is then used as a terminal cost or shaping function for MPC, thereby injecting long-horizon foresight into an otherwise short-horizon optimizer. 
These approaches share the same motivation as ours, i.e., mitigating the short-horizon bias of online optimization, but are orthogonal in how they inject long-term structure. Even with an accurate estimate of the infinite-horizon value function, MPC still requires an effective mechanism for \emph{searching} the control space. In contrast, our method focuses directly on improving this search by learning generative models over successful control sequences, thereby guiding the sampling distribution used by SPC. 

\section{Background}
\label{sec:background}
\subsection{Problem Formulation}
\label{sec:problem_formulation}
In this paper we consider optimal control problems in continuous action spaces. 
Given an initial state $\bm{x}_0=\bm{x}_{init}$, the objective is to determine a sequence of open-loop control actions $U_\tau=[\bm{u}_\tau, \bm{u}_{\tau+1}, \dots, \bm{u}_{\tau+T}]$ that minimizes a given cost function $\ell(\bm{x}, \bm{u})$ over a finite time horizon $T$:
\begin{align}\label{eq:problem_formulation}
\min_{\bm{u}_0, \bm{u}_1, \ldots, \bm{u}_T} \quad & L_f(\bm{x}_{T+1}) + \sum_{\tau=0}^{T} \ell(\bm{x}_\tau, \bm{u}_\tau) \tag{1a} \\
\text{s.t.} \quad & \bm{x}_{\tau+1} = f(\bm{x}_\tau, \bm{u}_\tau), \quad \tau = 0, \dots, T \tag{1b} \\
& \bm{x}_0 = \bm{x}_{\text{init}}, \tag{1c}
\end{align}
\vspace{-7pt}

where $\bm{x}_\tau \in \mathbb{R}^n$ and $\bm{u}_\tau \in \mathbb{R}^m$ are the state vector and the control input at time step $\tau$, respectively. The functions $\ell : \mathbb{R}^n \times \mathbb{R}^m \to \mathbb{R}$ and $L_f : \mathbb{R}^n \to \mathbb{R}$ represent the stage and terminal cost, respectively. We assume access to a simulator (e.g. MuJoCo \cite{todorov2012mujoco}) or a learned model to approximate the system dynamics $f : \mathbb{R}^n \times \mathbb{R}^m \to \mathbb{R}^n$.
For a more compact notation, we define a cost function $J : \mathbb{R}^{m \times T} \times \mathbb{R}^n \to \mathbb{R}$ that encapsulates both, costs and system dynamics, allowing us to write the problem as
\begin{equation}
\min_{U} \quad J(U; \bm{x}_{\text{init}}). \tag{2}
\end{equation}
Rather than deriving a single, globally optimal policy, MPC re-optimizes a local policy at each time step by simulating the system dynamics over a shorter receding horizon $H < T$. 

\subsection{Sampling-based MPC (SPC)} 
Contact-rich robot control tasks pose significant challenges due to non-convex cost functions and the nonlinear, often discontinuous nature of system dynamics. Sampling-based MPC addresses these issues by optimizing over a parameterized distribution $\pi_\phi(U)$ rather than directly computing the optimal control sequence.
We consider a generic SPC procedure in which, at each control step \( \tau \), the controller samples \( N \) control sequences \( \{U^{(i)}\}_{i=1}^N \) from the current distribution \( \pi_\phi \), simulates their outcomes from the current state estimate \( \hat{\bm{x}}_\tau \), and evaluates them using the cost function \( J(U^{(i)}; \hat{\bm{x}}_\tau) \). Based on these evaluations, the distribution parameters \( \phi \) are updated according to the chosen SPC algorithm. The executed control \( \bm{u}_\tau \) is typically the first element of the sampled control sequence \({U}_\tau \) or derived via spline interpolation across the optimized sequence.
We focus on diagonal Gaussian distributions of the form \( \pi_\phi(U) = \mathcal{N}(\bar{U}, \Sigma) \), as used in the Cross-Entropy Method (CEM)~\cite{rubinstein2004cross} and other SPC algorithms~\cite{howell2022predictive,williams2017model}. Here, \( \phi = (\bar{U}, \Sigma) \), and \( \Sigma = \mathrm{diag}(\bm{s}) \), with \( \bm{s} \) denoting a vector of variances.

\subsection{Generative Modeling: Flow-Matching}
While the above focuses on shaping a sampling distribution for SPC, generative modeling focuses on a different problem: produce a sample $x$ from a target distribution $p(x)$, which is typically unknown in closed form, but can be approximated by a dataset of samples $\mathcal{D}$. Among recent approaches to generative modeling, two closely related approaches have gained significant traction due to their ability to capture complex, multi-modal distributions: flow matching \cite{lipman2022flow} and diffusion models \cite{ho2020denoising}. The underlying concept is to learn a distribution over trajectories or transformations that maps a simple prior distribution to complex target data. 
In this work, we focus on flow-matching models and their conditional variant~\cite{tong2023improving}, as they offer superior inference speed compared to diffusion models. Flow matching aims to learn a time-dependent vector field \( v_\theta(\mathbf{x}, t) \) that transports samples from an easy-to-sample prior distribution \( p_0(\mathbf{x}) \) (e.g., a standard Gaussian) to a target data distribution \( p_1(\mathbf{x}) \). 
\section{Bootstrapping Sampling-based MPC with Generative Flow-Matching Models}
\label{sec:method}
In this section, we introduce our approach to bootstrapping SPC with generative models trained on open-loop control sequences collected from SPC itself. The core idea is to learn a generative model that approximates the distribution of successful control sequences conditioned on the current state and history. At test time, this model serves as a proposal distribution to guide and warm-start the sampling process in online SPC, improving sample efficiency and robustness.

\subsection{Data collection}
\label{sec:data_collection}
The offline data collection phase is central to our approach. It should provide control sequences (conditioned on the current state and history of states) likely to lead to task success. Since we aim to bootstrap SPC at test time, we generate this dataset directly from an SPC algorithm --- specifically CEM, as it is widely used and easy to tune for different tasks without the runtime constraints of online control. This allows for longer horizons and larger sample sizes to collect high-quality, non-myopic control sequences.

Given a task and associated cost function, we run CEM across multiple episodes, each with random state initializations and capped at a maximum number of MPC iterations.
During each episode, we record \( (\bm{x}_\tau, \bm{h}_\tau, U^*_\tau) \), where \( \bm{x}_\tau \) is the current state, \( \bm{h}_\tau \) encodes a fixed-window history of states, and \( U^*_\tau \) denotes the mean control sequence of the CEM distribution at time \( \tau \). An experiment is considered successful if the task is completed within the allowed time steps. We define \( \mathcal{I}_{\text{success}} \) as the index set of all successful experiment episodes and construct our training dataset as
\[
\mathcal{D} = \bigcup_{i \in \mathcal{I}_{\text{success}}} \{ (\bm{x}_\tau^{(i)}, \bm{h}_\tau^{(i)}, U_\tau^{*(i)}) \}_{\tau = 0}^{T_i},
\]
where \( T_i \) is the final time step of episode \( i \). This ensures that only control sequences from successful rollouts are used to train the generative model.

To reduce the complexity of the generative model while improving runtime efficiency and smoothness, each control sequence \( U_\tau \) is represented using \( K < H \) spline interpolation points over a planning horizon of \( H \) time steps, i.e. $U=[\bm{u}_0, \bm{u}_1, \dots, \bm{u}_K]$. We also employ \textit{i)} a progress-based heuristic to reset the variances during CEM to avoid early mode collapse, and \textit{ii)} action-level annealing \cite{xue2024full} that increases exploration, i.e., variances, for control points further into the horizon.

\subsection{Learning Control Sequence Proposal Distributions}
\label{sec:model_learning}
Once we have collected a task dataset of open-loop trajectories, we can train a flow-matching generative model to learn a time-varying state-conditional vector field $v_\theta(U, \bm{x}_\tau, \bm{h}_\tau, t)$ that pushes samples from the noise distribution $U_{t=0} \sim \mathcal{N}(0, I)$ to the target distribution $U_{t=1} \sim p_\theta(U \mid \bm{x}_\tau, \bm{h}_\tau)$, i.e. the distribution of control sequences that are likely to lead to successful task completion given the current state $\bm{x}_\tau$ and state history $\bm{h}_\tau$. For simplicity, we describe sampling from the generative model as sampling from the distribution $p_\theta(U \mid \bm{x}_\tau, \bm{h}_\tau)$. We refer to our method as generative predictive control (GPC), which leverages a learned distribution over control sequences conditioned on the task context. This distribution can be used in two distinct ways: \textit{i)} to sample control sequences from using a random shooting approach and evaluate the best based on value functions, or \textit{ii)} to update the sampling distribution of the SPC algorithm  (e.g., $\pi_\phi(U)$ in CEM) with samples from $p_\theta(U \mid \bm{x}_\tau, \bm{h}_\tau)$. We call the first approach \textit{GPC-Shoot} and the second approach \textit{GPC-CEM}.

\subsection{\textbf{GPC-CEM:} Bootstrapping SPC with Flow-Matching}
\begin{algorithm}[t]
\caption{GPC-CEM}
\label{alg:cem_flow}
\SetKwInput{Input}{Input}
\Input{\small
 Current state $\bm{x}_\tau$, history $\bm{h}_\tau$ \\
\hspace{0.95cm} Sampling distribution $\pi_\phi(U) = \mathcal{N}(\bar{U}, \Sigma)$ \\
\hspace{0.95cm} Flow model $p_\theta(U \mid \bm{x}_\tau, \bm{h}_\tau)$ \\
\hspace{0.95cm} Number of rollouts $N = N_{\text{CEM}} + N_{\text{Flow}}$ \\
\hspace{0.95cm} Number of elites $N_{\text{elite}}$ \\
\hspace{0.95cm} State estimator $\hat{\bm{x}}(\tau)$
}
\While{planning}{
    $\bm{x}_0 \gets \hat{\bm{x}}(\tau)$ \tcp*[r]{Get current state estimate}

    Sample $N_{\text{CEM}}$ trajectories from CEM: \\
    \hspace{0.3cm} $\{ U^{(i)} \}_{i=1}^{N_{\text{CEM}}} \sim \pi_\phi(U)$\;

    Sample $N_{\text{Flow}}$ trajectories from flow model: \\
    \hspace{0.3cm} $\{ U^{(j)} \}_{j=1}^{N_{\text{Flow}}} \sim p_\theta(U \mid \bm{x}_\tau, \bm{h}_\tau)$\;

    Compute $\{ J^{(k)} \gets J(U^{(k)}; \bm{x}_0) \}_{k=1}^{N}$ \tcp*[r]{parallel rollouts}

    Select top $N_{\text{elite}}$ elite trajectories with lowest cost: \\
    \hspace{0.3cm} $\{ U^{(k)} \}_{k=1}^{N_{\text{elite}}} \gets$ elite set\;

    Update $\pi_\phi(U)$ using elite statistics: \\
    \hspace{0.3cm} $U^* \gets U^{(k^*)}, \quad k^* = \arg\min_k J^{(k)}$ 

    \hspace{0.3cm} $\bar{U} \gets \texttt{shift}(U^*, \tau)$ \tcp*[r]{shift mean forward}
    \hspace{0.3cm} $\Sigma \leftarrow \mathrm{diag}\left(\mathrm{Var}(\{ U^{(k)} \}_{k=1}^{N_{\text{elite}}})\right)$\;

    Execute control $\bm{u}_\tau \gets \texttt{get\_action}(U^*, \tau)$\;

    Update state history $\bm{h}_{\tau+1} \leftarrow \texttt{roll}( \bm{h}_\tau, \bm{x}_\tau )$\;
}
\vspace{-5pt}
\end{algorithm}
Trained on a finite set of open-loop control sequences, the generative proposal distribution \( p_\theta(U \mid \bm{x}_\tau, \bm{h}_\tau) \) inherits the generalization limitations of behavior cloning and model-based RL. This sensitivity to distributional shifts is something we also observe in our experiments with GPC-Shoot, where it manifests as degraded sample quality within regions underrepresented in the training data. In contrast, SPC adapts its sampling distribution online and becomes more robust in unseen situations, but remains myopic and computationally expensive. To balance these limitations, we bootstrap SPC with a flow-matching generative model that learns the dataset control distribution while preserving the adaptability of online sampling.
%
%
We summarize our approach in Algorithm~\ref{alg:cem_flow}.
At each control step, the algorithm begins by estimating the current state and shifting the current mean of the CEM sampling distribution forward in time. 
The key idea in GPC-CEM is to augment Gaussian CEM sampling with proposals from the generative model \( p_\theta \) trained offline on control sequences that led to task success. The $N_{elite}$ proposals with the lowest-cost rollouts are used to update the CEM distribution's \( \pi_\phi(U) \)  mean (the time-shifted lowest-cost proposal) and variance. Unlike standard CEM, the executed control is the best-performing candidate instead of the mean of the $N_{elite}$ proposals. This allows GPC-CEM to better exploit multimodal proposals from the generative model rather than collapsing them to a single modality, efficiently guiding exploration while maintaining the adaptability of online optimization.
\subsection{Application to Loco-Manipulation}
We apply our bootstrapped SPC framework to non-prehensile object pushing with a Spot quadruped robot to demonstrate versatile loco-manipulation skills. With 19 positional degrees of freedom (DoF), planning for this robot is computationally expensive and typically demands large sample sizes.
To simplify our sampling process, we disentangle low-level locomotion control based on the work of~\cite{zhu2025versatile} and sample only in the high-level task action space.
The high-level action space includes 9 DoF (3 for the torso, 6 for the arm)\footnote{We exclude the gripper DoF from the high-level action space for non-prehensile manipulation, but this and additional DoFs (e.g., torso height, pitch, roll) could be included without retraining the low-level policy.}
and is mapped to the low-level commands by a pre-trained locomotion policy that ensures balance and stability while tracking high-level inputs.
The low-level locomotion policy is fixed throughout task planning and execution. In addition to a lower-dimensional action space, this hierarchical control structure naturally provides more robustness to the low-level control.
As a result, we do not need to explicitly enforce strong smoothness or temporal consistency constraints in the high-level action space used for flow matching.


\section{Experimental Results}
\label{sec:results}

We evaluate our proposed GPC framework across simulated and real-world continuous control tasks involving contact-rich, non-prehensile manipulation. Specifically, we benchmark performance on \textit{i)} the well-known Push-T task with a 2-DoF circular robot, and \textit{ii)} the loco-manipulation task introduced above. 
To guide our evaluation, we aim to answer the following key questions:
\textit{i)} How well does the learned generative model approximate the action proposal distribution captured by open-loop sampling-based MPC? \textit{ii)} Does bootstrapping online MPC with a learned proposal distribution improve \textit{task performance} and \textit{generalization to task variations} under constrained computational budgets?

\begin{figure}[t]
\centering
\includegraphics[width=0.7\linewidth]{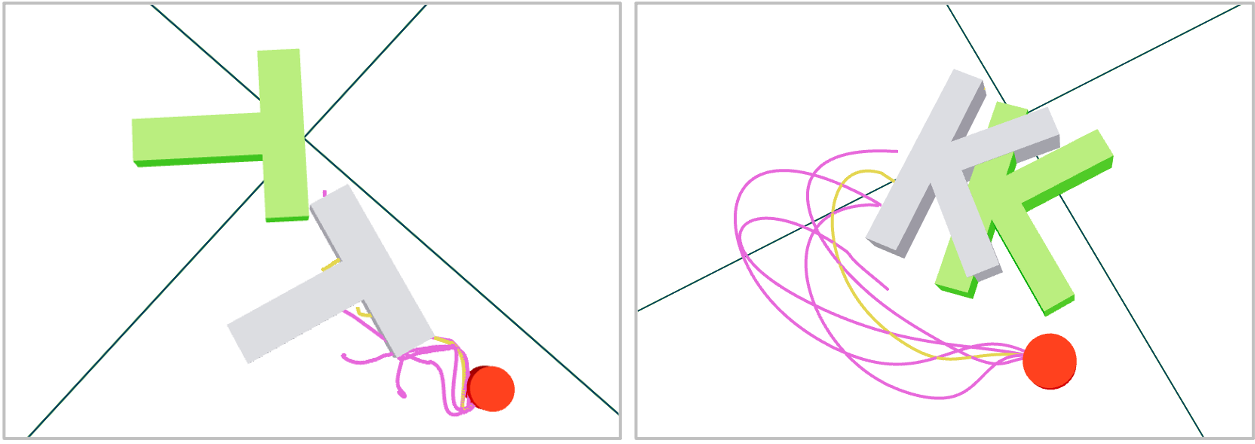}
\caption{\textbf{Push-T Task overview.} \textit{Left:} Original Push-T task \cite{chi2023diffusion}, where a circular robot is required to push a T-shaped block into a target pose, shown in green. \textit{Right:} Modified task with a K- instead of T-block at the bottom.}  
\label{fig:task_overview}
\vspace{-10pt}
\end{figure}

We use conditional flow matching to train a generative model over SPC control sequences with a Multi-Layer Perceptron (MLP) as the underlying architecture. We also baseline the flow model against a CVAE trained on the same data.
We consider both direct sampling from the learned model (\emph{GPC-Shoot}) and the bootstrapped version that combines it with CEM-based online planning (\emph{GPC-CEM}).
We compare our approach to standard CEM, model predictive path integral control (MPPI); \cite{williams2017model}; and DialMPC~\cite{xue2024full}, a more recent SPC approach building on MPPI.
We only compare against DialMPC and GPC-Shoot with a CVAE on Push-T; the former's inner optimization loop makes it unsuitable for our real-time loco-manipulation examples, while the latter proved inferior to flow matching.
All methods are implemented in Python using \texttt{judo}~\cite{li2025_judo} as a unified interface for defining custom tasks and controllers. For each task, we evaluate all methods with the same number of rollouts per iteration, control frequency, and respective cost function. In addition, we report results for the GPC-methods across three different model seeds to account for the stochasticity during training. We set the CEM-sample ratio in GPC-CEM, i.e. $N_{\mathrm{CEM}} /N$, to 0.5 for both tasks.

\begin{table}[t]
    \centering
    \renewcommand{\arraystretch}{1.02}
        \centering
        \small
        \caption{Simulation Results for the Push-T task, including a horizon ablation and a task variation with a K- instead of a T-block.}

        \resizebox{\columnwidth}{!}{
        \begin{tabular}{lccc}
            \toprule
            \rowcolor{gray!15}
            \multicolumn{4}{c}{\textit{Control frequency:} 10 Hz \quad \textit{Time step ($\Delta t$):} 0.01 s \quad \textit{Rollouts:} 32} \\
            \addlinespace[0.5ex]  
            & \textbf{Success rate} & \textbf{Number of steps} & \textbf{CEM} \\
            & ($\uparrow$) & (success only, ($\downarrow$)) & \textbf{sample ratio} \\
            \midrule
            \multicolumn{4}{l}{\textit{Base Task: Push-T}} \\
            \midrule
            CEM Baseline & 0.85 (0.83, 0.88) & 1037.57 $\pm$ 526.40 & -- \\
            MPPI Baseline & 0.62 (0.59, 0.65) & 1634.15 $\pm$ 492.42 & -- \\
            Dial-MPC~\cite{xue2024full} & 0.86 (0.83, 0.88) & 1197.07 $\pm$ 461.16 & -- \\
            \arrayrulecolor{gray!50}
            \midrule
            \arrayrulecolor{black}
            GPC-Shoot (CVAE) & 0.970 (0.951, 0.982) & 985.31 $\pm$ 475.07 & -- \\
            GPC-Shoot (2) & 0.718 (0.702, 0.734) & 1277.51 $\pm$ 572.80 & -- \\
            GPC-Shoot (10) & 0.992 (0.988, 0.995) & 608.58 $\pm$ 291.48 & -- \\
            GPC-CEM (2) & 0.980 (0.980, 0.985) & 932.40 $\pm$ 449.95 & 0.33 $\pm$ 0.11 \\
            GPC-CEM (10) & \textbf{0.998} (0.996, 0.999) & \textbf{591.11} $\pm$ 267.20 & 0.33 $\pm$ 0.10 \\
            \midrule
            \multicolumn{4}{l}{\textit{Horizon Ablation: using 1 secs. instead of 3 secs. at inference time}} \\
            \midrule
            CEM Baseline & 0.78 (0.76, 0.81) & 1093.50 $\pm$ 523.09 & -- \\
            MPPI Baseline & 0.68 (0.65, 0.71) & 1365.65 $\pm$ 463.48 & -- \\
            Dial-MPC~\cite{xue2024full} & 0.84 (0.81, 0.86) & 945.76 $\pm$ 403.05 & -- \\
            \arrayrulecolor{gray!50}
            \midrule
            \arrayrulecolor{black}
            GPC-Shoot (CVAE) & 0.71 (0.67, 0.75) & 1209.23 $\pm$ 581.55 & -- \\
            GPC-Shoot (10) & 0.84 (0.83, 0.85) & 978.96 $\pm$ 543.72 & -- \\
            GPC-CEM (10) & \textbf{0.96} (0.95, 0.97) & \textbf{890.42} $\pm$ 466.94 & 0.42 $\pm$ 0.08 \\
            \midrule
            \multicolumn{4}{l}{\textit{Task Variation: Push-K}} \\
            \midrule
            CEM Baseline & 0.55 (0.52, 0.58) & 1143.21 $\pm$ 565.90 & -- \\
            MPPI Baseline & 0.34 (0.31, 0.36) & 1707.92 $\pm$ 501.53 & -- \\
            Dial-MPC~\cite{xue2024full} & 0.56 (0.52, 0.59) & 1333.13 $\pm$ 484.97 & -- \\
            \arrayrulecolor{gray!50}
            \midrule
            \arrayrulecolor{black}
            GPC-Shoot (CVAE) & 0.51 (0.46, 0.55) & 1055.39 $\pm$ 539.90 & -- \\
            GPC-Shoot (10) & 0.89 (0.88, 0.90) & 1015.15 $\pm$ 527.04 & -- \\
            GPC-CEM (10) & \textbf{0.96} (0.95, 0.96) & \textbf{887.77} $\pm$ 482.02 & 0.41 $\pm$ 0.10 \\
            \bottomrule
        \end{tabular}}
        \label{tab:pusht_pushk_metrics_combined}
\vspace{-8pt}
\end{table}


\subsection{Push-T Task}
This task requires a 2-DoF circular robot to push a T-shaped block to a specified goal pose. Due to its sparse rewards and multi-modality, it serves as a popular benchmark for evaluating generative control policies. We also evaluate the adaptability of GPC to unseen task variations by running it on a variant using a K-shaped block (Push-K) with different object dynamics. In this setting, we reuse the generative model trained on Push-T to bootstrap SPC for Push-K without retraining, showcasing GPC’s ability to generalize across task variations.
Table~\ref{tab:pusht_pushk_metrics_combined} summarizes the simulation results. We report success rates with Wilson 95\% confidence intervals and average completion times (for successful runs) with respective standard deviations based on 1000 trials per method. Success is defined as achieving at least 90\% coverage of the targetxx pose within 2500 time steps (0.01\,s each). For GPC-CEM, we additionally report the average CEM sample ratio and standard deviation, indicating how often samples were selected from the CEM distribution over the learned proposal distribution.

Both CEM and DIAL-MPC~\cite{xue2024full} achieve high success rates~($\geq85$\%), demonstrating the strength of sampling-based methods. DIAL-MPC's gains over CEM are marginal and come at higher computational cost, so we use CEM for data collection. MPPI achieves 62\% success, though better tuning may close this gap.
Flow-based GPC-Shoot improves with more denoising steps (indicated in parentheses): 99\% success with 10 steps vs.\ 71\% with 2, while CVAE-based GPC-Shoot achieves 97\% success. Similarly, GPC-CEM with 10 steps not only achieves 99.8\% success but also reduces completion time by nearly 50\% compared to 2 steps. Notably, GPC-CEM remains robust under reduced horizons (1s vs.\ 3s), maintaining 96\% success versus 78\% for CEM. This suggests that our method enables non-myopic planning under tighter computational constraints.
In the Push-K generalization task, GPC-CEM again outperforms all baselines, achieving 96\% success compared to 55\% for CEM, indicating strong transferability of the learned proposal distribution. This is a notable insight, as both CVAE-based GPC-Shoot and CEM performance drops significantly without any changes to the cost function, highlighting that the flow-based learned distribution captures structural priors useful across tasks.

\begin{figure}[t]
    \centering

    \begin{minipage}{0.49\columnwidth}
        \centering
        \includegraphics[width=\linewidth]{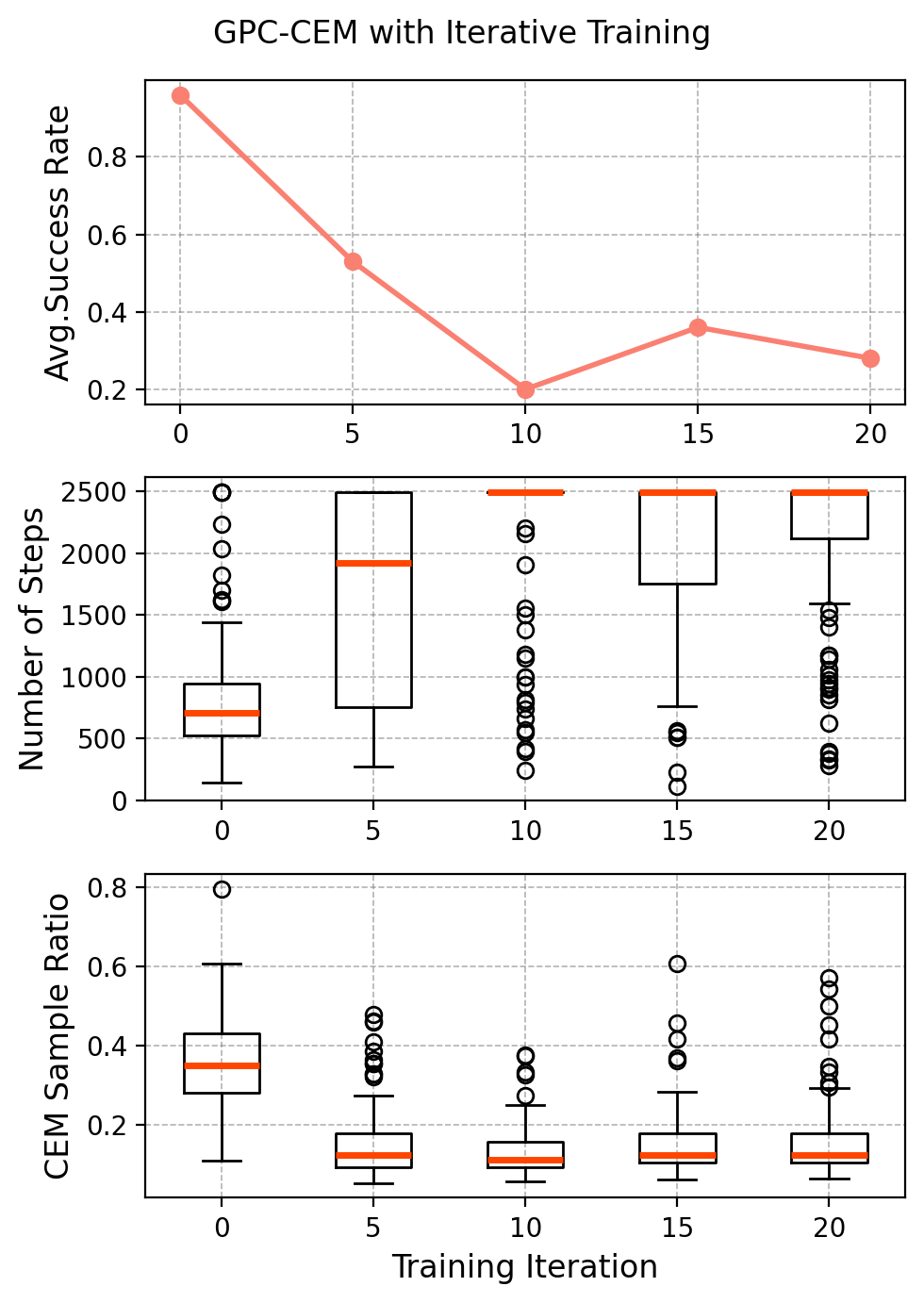}
    \end{minipage}
    \hfill
    \begin{minipage}{0.49\columnwidth}
        \centering
        \includegraphics[width=\linewidth]{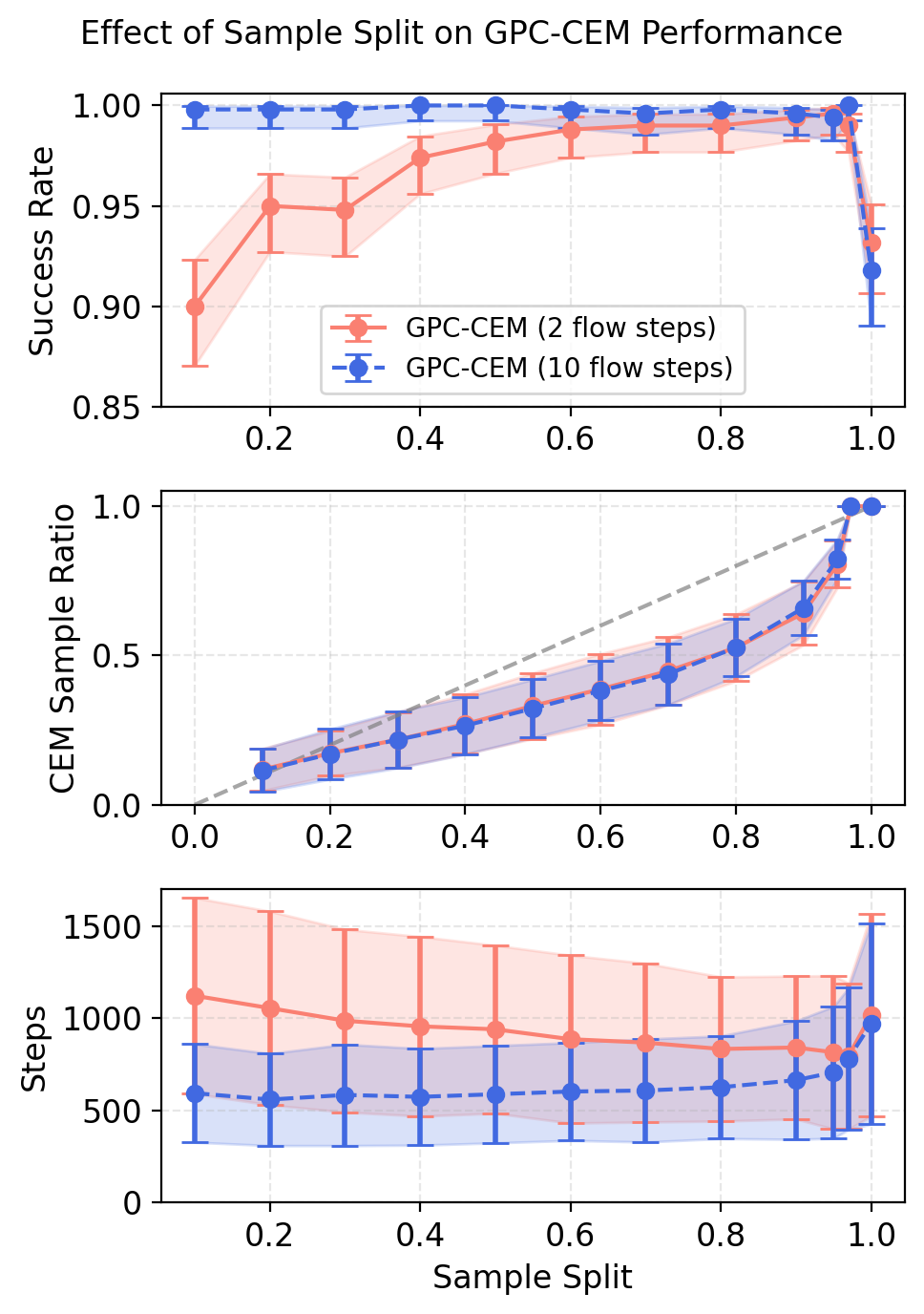}
    \end{minipage}

    \caption{\textbf{Push-T Simulation Studies.} 
        \emph{Left:} Evaluation of intermediate models from the iterative training procedure of~\cite{kurtz2025generative}. 
        \emph{Right:} Ablation study on sample split.
    }
    \label{fig:iterative_training_comparison}
    \vspace{-15pt}
\end{figure}
\paragraph*{Comparison to Iterative Training Procedure \cite{kurtz2025generative}}
We acknowledge the conceptual similarity between our approach and that of Kurtz et al.~\cite{kurtz2025generative}, who also combine generative modeling with SPC. However, their method adopts an iterative training procedure that alternates between data collection and model updates, similar to expert iteration in reinforcement learning~\cite{anthony2017thinking}. This setup is motivated by the assumption that SPC data is too noisy to directly train a generative model; hence, each data collection iteration bootstraps SPC with a partially trained flow-matching model to improve the subsequent training distribution.
In our experiments on the Push-T task, however, this iterative procedure did not improve performance (see Fig.~\ref{fig:iterative_training_comparison}). In fact, success rates decline over training iterations. In a qualitative analysis, we find that the resulting policies tend to collapse to small, random movements that fail to complete the task. We interpret this as the iterative training procedure gradually diminishing the multi-modality of the learned proposal distribution. Consistent with this interpretation, we also observe a decreasing CEM sample ratio over training iterations, suggesting that the learned proposal distribution converges to mimic the CEM sampling distribution and reduces their complementarity. 
%
In contrast, our method trains a generative model directly on open-loop control sequences from SPC, without requiring iterative retraining. Despite the noisy data, our model achieves up to 99.2\% success (GPC-Shoot with 10 denoising steps) and already reaches 96\% when bootstrapped with CEM after a single training round.

\paragraph*{Ablations}
We conduct three ablation studies to further analyze the performance of GPC-CEM and GPC-Shoot on the simulated Push-T task.
\textbf{Ablation 1} measures the effect of the sample split between samples drawn from the CEM distribution and the learned proposal distribution. We vary the sample split $N_{\mathrm{CEM}} /N$ from 0.1 to 1.0 (only CEM). As shown in Fig.~\ref{fig:iterative_training_comparison}, we find that increasing the number of CEM samples improves or maintains the success rate until exclusively using CEM samples, where performance drops significantly. This highlights that the learned proposal distribution contributes valuable samples that both complement and strengthen the CEM distribution.
We further observe that the empirical CEM sample ratio grows sublinearly with respect to the specified split, indicating that even a relatively small fraction of learned proposal samples disproportionately contribute to the overall sampling process.
\textbf{Ablation 2} analyzes the impact of the dataset size used to train the flow-matching model. To assess how many demonstrations are needed to train an effective proposal model, we vary the dataset size from 100 to 1000 MPC rollouts\footnote{Each rollout corresponds to a single trajectory of $H/\Delta t$ steps.}. Performance improves rapidly with data and surpasses 90\% success with only 200 rollouts, after which performance gains steadily saturate. This shows that the proposal model can be trained in a sample efficient manner requiring just a few hundred trajectories, whereas reinforcement learning methods typically need orders of magnitude more interaction to achieve comparable performance.
\textbf{Ablation 3} assesses model architecture by comparing flow model performance against a conditional variational auto-encoder (CVAE) using GPC-Shoot. Table~\ref{tab:pusht_pushk_metrics_combined} shows the ten-step flow model outperformed the CVAE for each task and variation with a similar number of steps. Along with its degraded performance on the Push-K task, the CVAE's struggle to adjust to a different horizon length was characteristic of its documented difficulties with multi-modal generalization and domain adaptation \cite{daunhawer2022on}. Due to its higher performance and robust generalization, we selected flow matching as our generative model. 

\subsection{Spot Loco-Manipulation}
\begin{figure}[t]
    \centering
    \includegraphics[width=0.97\linewidth]{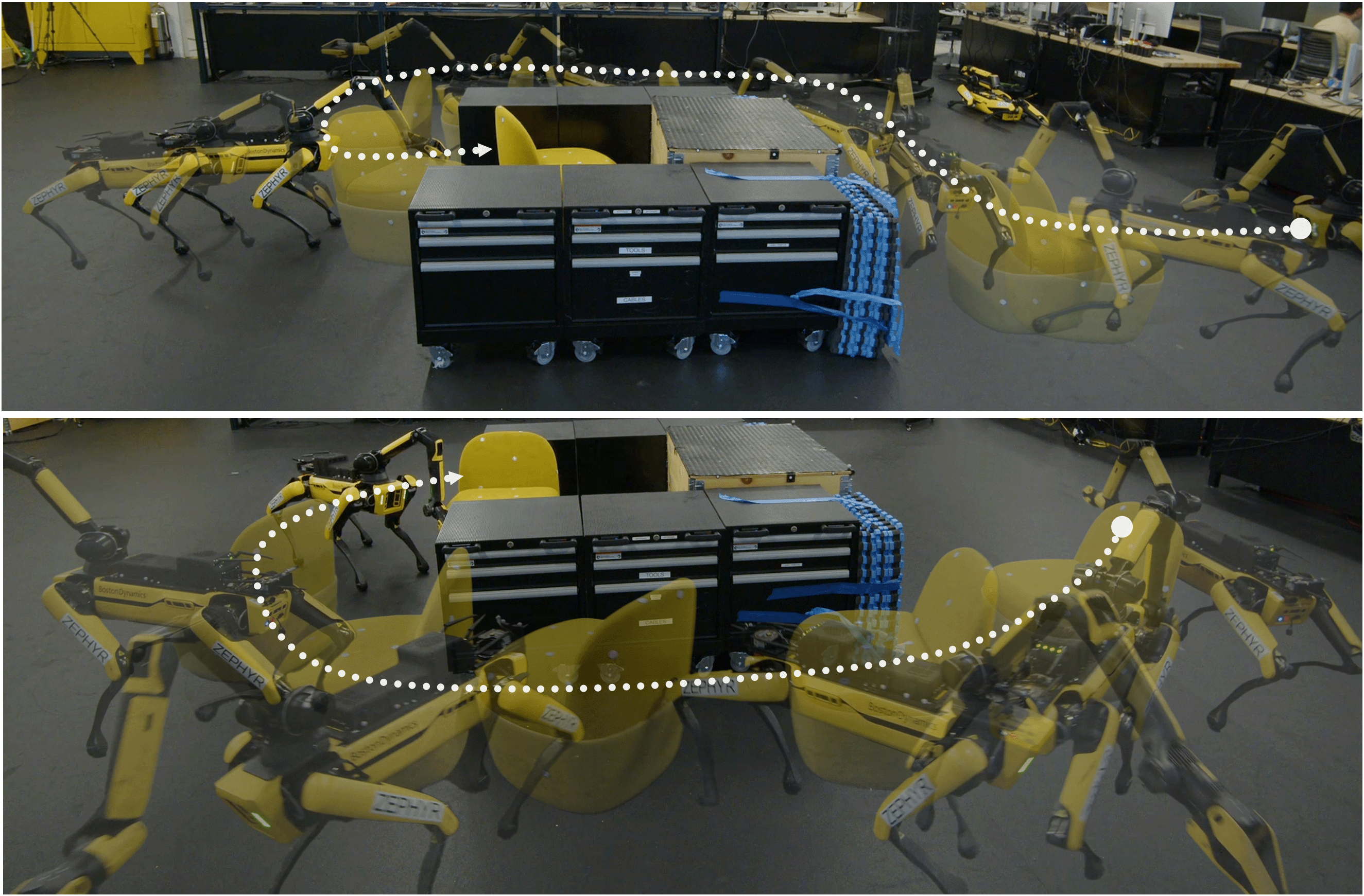}
    \caption{\textbf{Spot Loco-Manipulation Experiment:} Qualitative examples of two successful Spot loco-manipulation task runs in the real world with GPC-CEM. Both images show a overlay of several snapshots of the trajectories of the robot and the chair. }
\vspace{-10pt}
\label{fig:spot_exp_qualitative}
\end{figure}

In this task, Spot must push a chair to a goal pose located behind a C-shaped obstacle (Fig.~\ref{fig:spot_exp_qualitative}). The robot and chair are always initialized randomly on the opposite side of the obstacle, creating a local minimum that requires navigating around it to succeed. The task is further complicated by the high-dimensional action space and contact dynamics. Solving this with SPC requires long horizons and large sample sizes, both of which increase computational cost. A trial is considered successful if the chair's position error is below 0.15\,m and its yaw is within 50 degrees of the target.

\begin{table}[t]
    \centering
    \caption{Simulation Results for the Spot loco-manipulation task, including horizon ablation and task variation with additional obstacle avoidance cost at runtime.}
    \scriptsize
    \setlength{\tabcolsep}{2.8pt}
    \renewcommand{\arraystretch}{1.05}
    \begin{tabular}{lcccc}
        \toprule
        \rowcolor{gray!15}
        \multicolumn{4}{c}{\textit{Control frequency:} 5 Hz \quad \textit{Time step ($\Delta t$):} 0.02 s \quad \textit{Rollouts:} 32} \\
        \addlinespace[0.5ex]  
        & \textbf{Succ. rate} & \textbf{Number of steps} &  \textbf{CEM} \\
        & ($\uparrow$) & (success only, ($\downarrow$)) & \textbf{sample ratio}\\
        \midrule
        \multicolumn{4}{l}{\textit{Base Task: Spot Loco-Manipulation}} \\
        \midrule
        CEM Baseline & 0.33 (0.28, 0.39) & 1452.1 $\pm$ 448.2 & -- \\
        MPPI Baseline & 0.57 (0.51, 0.62) & 1096.3 $\pm$ 360.6 & -- \\
        \arrayrulecolor{gray!50}
        \midrule
        \arrayrulecolor{black}
        GPC-Shoot (2) & 0.18 (0.14, 0.23)& 1544.6 $\pm$ 495.2 & -- \\
        GPC-Shoot (10) & 0.22 (0.18, 0.27) & 1454.2 $\pm$ 511.5 & -- \\
        GPC-CEM (2) & \textbf{0.83} (0.78, 0.87) & 1125.9 $\pm$ 430.6 & 0.69 $\pm$ 0.10 \\
        GPC-CEM (10) & 0.79 (0.74, 0.83) & \textbf{1073.9 $\pm$ 367.4} & 0.66 $\pm$ 0.11 \\
        \midrule
        \multicolumn{4}{l}{\textit{Horizon Ablation: using 3 secs. instead of 4 secs. at inference time}} \\
        \midrule
        CEM Baseline & 0.26 (0.21, 0.31) & 1494.6 $\pm$ 491.9 & -- \\
        MPPI Baseline & 0.29 (0.24, 0.34) & 1177.3 $\pm$ 465.3 & -- \\
        \arrayrulecolor{gray!50}
        \midrule
        \arrayrulecolor{black}
        GPC-Shoot (2) & 0.22 (0.18, 0.27) & 1489.3 $\pm$ 498.7 & -- \\
        GPC-CEM (2) & \textbf{0.60} (0.54, 0.65) & 1135.3 $\pm$ 487.9 & 0.71 $\pm$ 0.08 \\
        \midrule
        \multicolumn{4}{l}{\textit{Task Variation: Spot Loco-Manipulation with Obstacle Avoidance}} \\
        \midrule
        CEM Baseline & 0.27 (0.22, 0.32) & 1692.3 $\pm$ 439.7 & -- \\
        MPPI Baseline & 0.30 (0.25, 0.35) & 1634.2 $\pm$ 473.9 & -- \\
        \arrayrulecolor{gray!50}
        \midrule
        \arrayrulecolor{black}
        GPC-Shoot (2) & 0.03 (0.02, 0.06) & 1764.5 $\pm$ 353.2 & -- \\
        GPC-CEM (2) & \textbf{0.56} (0.50, 0.62) & \textbf{1421.1 $\pm$ 481.8} & 0.74 $\pm$ 0.08 \\
        \bottomrule
    \end{tabular}
    \label{tab:base_and_variant_metrics}
    \vspace{-10pt}
\end{table}

\paragraph*{Simulation} 
We first evaluate the task in simulation to enable larger-scale testing. Results are summarized in Table~\ref{tab:base_and_variant_metrics}. Each baseline is run for 100 trials, and GPC methods are evaluated with 3 model seeds (100 trials each). We omit DIAL-MPC due to its inner-loop optimization being too slow for real-time use in this task.
As in Push-T, we report success rate, average completion steps (for successful runs), and CEM sample ratio. GPC-CEM outperforms all baselines, achieving up to 83\% success with fewer executed steps. In contrast, CEM alone reaches only 33\%, often failing due to limited horizon and sample budget. MPPI performs slightly better but remains unreliable under real-time constraints.
GPC-CEM remains robust under reduced planning horizons (3 vs.\ 4 seconds), maintaining 60\% success while baseline performance degrades. This reinforces that learned proposals can enhance planning in resource-limited settings.
Interestingly, fewer denoising steps (2 vs.\ 10) yield better performance in this task for both GPC-Shoot and GPC-CEM. We attribute this to reduced sample diversity at higher step counts, which impairs exploration in tasks with deceptive local minima. We also observe higher CEM sample ratios in this task compared to Push-T, indicating that the learned model alone (GPC-Shoot) is less accurate. Instead, it is most effective when used to augment CEM, highlighting the value of integrating learned proposals into online optimization rather than relying on them directly.
Finally, we evaluate a task variant with an added obstacle avoidance cost to prevent collisions with the C-shaped obstacle. This is omitted from the base task to avoid biasing the MPC methods, but is essential for real-world deployment. In this setting, GPC-CEM still leads with a 56\% success rate, outperforming MPPI (30\%) and CEM (27\%).

\paragraph*{Real-World}
We evaluate GPC-CEM and CEM on hardware including the obstacle avoidance cost in both cases. 
We rely on a Motion Capture system to track the object state.
GPC-CEM is run with 2 denoising steps for 20 trials, while CEM is limited to 10 trials due to frequent damaging failures to the robot (e.g. repeated collisions with the obstacle as it fails to navigate around). \textbf{GPC-CEM} achieves a \textbf{60\% success rate (12/20)}, while \textbf{CEM} succeeds in only \textbf{10\% of trials (1/10)}. Qualitative examples for GPC-CEM are shown in Fig.~\ref{fig:spot_exp_qualitative}; all other runs are included in the supplementary video\footnote{\url{https://youtu.be/lKCGjjddv1E}}.
CEM failures consistently result in the local minimum caused by the C-shaped obstacle, as it lacks the guidance from the generative model to sample motions that navigate around it. GPC-CEM only encounters this failure in 4 of 20 trials. The remaining failures stem from two causes: (1) pushing the chair beyond the workspace due to the lack of workspace constraints in the cost, and (2) discrepancies between simulated and real chair behavior, especially assumptions about friction and contact such as when the chair's wheels can roll. 
\paragraph*{Computation Time}  
We find that the policy rollout accounts for over 90\% of the total compute time and becomes the primary computation bottleneck, limiting the overall control frequency to 5\,Hz. This overhead is primarily due to collision handling and contact dynamics in the physics engine.





\section{Limitations and Future Work}
Our framework does not explicitly address sim-to-real transfer, leaving it vulnerable to discrepancies between simulated and real-world dynamics. However, since it relies on offline data collection, it can be trained on domain-randomized data to improve robustness to variations in dynamics, actuator behavior, and sensor noise~\cite{kurtz2025generative}. In addition, learning proposal distributions from a combination of simulated and real-world data could further enhance transferability and performance during hardware deployment.
In this work, all experiments consider fixed goals in a world frame. We plan to extend our work to variable goals by transforming our data into goal-centric representations. The current system also does not use GPU acceleration for simulation or proposal inference, but this could be addressed in future work to enable faster online planning and data collection (particularly with larger sample sizes).
Finally, the proposed approach is limited to state-based policies but can be distilled to vision-based policies by learning from observations collected while executing the state-based policy in the real world or simulation. Future work can explore how to integrate vision-based action proposal distributions with fast vision-based dynamics models for online predictive control~\cite{qi2025strengthening}.
Although our offline data collection already captures long-horizon structure by using extended SPC horizons, future work could also incorporate learned infinite-horizon value functions. Such value estimates would provide an additional source of global, task-level guidance, while our generative priors would continue to shape and improve the search over control sequences during online optimization.

\section{Conclusion}
We presented GPC-CEM, a generative predictive control (GPC) framework that bootstraps sampling-based MPC (SPC) with conditional flow-matching trained on open-loop SPC control sequences. Our approach demonstrates that meaningful proposal distributions can be learned directly from noisy SPC data, without expert supervision or iterative refinement.
We evaluated our method in two challenging settings; a simulated pushing benchmark and a real-world quadruped loco-manipulation task; and showed that it significantly improves sample efficiency and robustness. GPC-CEM achieves high success rates, remains effective under reduced planning horizons, and generalizes to task variations which introduce out-of-distribution conditions. These results highlight the effectiveness of integrating learned generative models into online optimization loops for efficient and adaptable real-time robot control.




\bibliographystyle{IEEEtran}
\bibliography{refs}

\section*{APPENDIX}
\subsection{Implementation Details}
We describe the implementation of our generative model and sampling-based MPC algorithm, used for both offline data collection and online control. 

\paragraph*{Data Collection}
We collect training data using a sampling-based MPC controller in simulation for a maximum of 2500 time steps per episodes in both tasks, which generates open-loop control sequences. For Push-T, trajectories are cubic splines with 4 control points; for Spot, they consist of 4 linearly interpolated waypoints. All data is represented in the world frame, not relative to the robot. Although we tested robot-centric representations, they did not yield performance improvements.
For Spot, the manipulated object is modeled as a single free rigid body with empirically tuned mass, inertia, and friction, using simplified collision geometry and low base friction to approximate rolling behavior.
We collected 67,667 Push-T sequences from 1,000 successful episodes and 211,832 Spot sequences from 1,700 episodes. For evaluation, we use fixed sets of randomly sampled initial states for each task, shared across all methods and runs to ensure consistency.


\paragraph*{Training Details}
We implemented our baseline CVAE and conditional flow-matching models as MLPs trained with a batch size of 40,000 and the Adam optimizer with a learning rate 0.0001, cosine annealing schedule, 500 warmup steps over 1,000 epochs. The model predicts 4 control points conditioned on the current robot and object state and the previous replanning state (history length = 1). Orientations are represented using sine-cosine encodings of yaw angles. Although Spot expects velocity commands, we predict absolute positions and convert them to velocities via finite differences during online control.
Weights were tuned empirically.
\paragraph*{Training Details}
We implemented our baseline CVAE and conditional flow-matching models as MLPs trained with a batch size of 40,000 and the Adam optimizer with a learning rate 0.0001, cosine annealing schedule, 500 warmup steps over 1,000 epochs. The model predicts 4 control points conditioned on the current robot and object state and the previous replanning state (history length = 1). Orientations are represented using sine-cosine encodings of yaw angles. Although Spot expects velocity commands, we predict absolute positions and convert them to velocities via finite differences during online control.
Weights were tuned empirically.
\paragraph*{Cost Functions}
Both tasks use a weighted sum of costs computed over the full-resolution control sequence (0.01s for Push-T, 0.02s for Spot). The cost components include \textit{robot–object proximity} (L2 distance between the robot and object, penalizing both torso and end-effector distances for Spot), a \textit{velocity penalty} given by the L2 norm of robot joint velocities, \textit{goal reaching} terms measuring L2 distance and angle difference between the object and goal with an additional progress penalty, \textit{joint limit penalties} for exceeding arm joint limits on Spot (leg joints handled by the low-level policy), a \textit{fall penalty} applied when the Spot torso height drops below a threshold, and an \textit{object tipping penalty} when the chair’s z-axis deviates from vertical.

\end{document}